\documentclass[sigconf]{acmart}
\AtBeginDocument{%
  }

\renewcommand\footnotetextcopyrightpermission[1]{}

\setcopyright{acmlicensed}
\copyrightyear{2018}
\acmYear{2018}
\acmDOI{XXXXXXX.XXXXXXX}
\acmConference[Conference acronym 'XX]{ACM Multimedia}{June 03--05,
  2018}{Woodstock, NY}
\acmISBN{978-1-4503-XXXX-X/2018/06}

\usepackage{hyperref}
\usepackage[table]{xcolor}

\usepackage{adjustbox}

\usepackage{orcidlink}

\usepackage{multirow}
\usepackage{xspace}

\usepackage{caption}
\usepackage{svg}

\definecolor{lightpurple}{RGB}{242, 242, 255}
\newcommand{\modelname}{D3O\xspace}

\newcommand{\eg}{\emph{e.g.}\xspace}

\begin{document}

\title{D3O: Dynamic Distribution Distillation for Ordinal Regression}

\author{Chunlai Dong}
\affiliation{%
  \institution{Zhejiang University}
  \city{Hangzhou}
  \country{China}}
\email{dongcl@zju.edu.cn}

\author{Yaojun Hu}
\affiliation{%
  \institution{Zhejiang University}
  \city{Hangzhou}
  \country{China}}
\email{yaojunhu@zju.edu.cn}

\author{Yuyang Xu}
\affiliation{%
  \institution{Zhejiang University}
  \city{Hangzhou}
  \country{China}}
\email{xuyuyang@zju.edu.cn}

\author{Haochao Ying}
\authornote{Haochao Ying and Jian Wu are the corresponding authors.}
\affiliation{%
  \institution{Zhejiang University}
  \city{Hangzhou}
  \country{China}}
\email{haochaoying@zju.edu.cn}

\author{Jian Wu}
\authornotemark[1]
\affiliation{%
  \institution{Zhejiang University}
  \city{Hangzhou}
  \country{China}}
\email{wujian2000@zju.edu.cn}

\renewcommand{\shortauthors}{Dong et al.}


\begin{abstract}
Ordinal regression is widely used in scenarios where labels are discrete yet inherently ordered. 
In practice, however, ordinal labels are often obtained by discretizing underlying continuous semantics through subjective human judgment, resulting in ambiguous class boundaries and annotation noise. 
This poses a fundamental challenge to existing methods that rely on static supervision, as predefined labels may impose globally rigid and even biased ordering constraints throughout training.
To address this limitation, we propose \modelname, a dynamic distribution distillation framework for ordinal regression. Beyond relying solely on predefined ordinal labels, \modelname introduces a dynamic evolution of label distributions via self-distillation.
Specifically, we introduce a contrastive ordinal-aware label enhancement module that leverages vision–language alignment to recover ordinal label distributions capturing both inter-class ambiguity and instance-level uncertainty. These recovered distributions serve as dynamic supervision signals that are iteratively refined throughout training.
Furthermore, we design a CDF-based cross-layer interaction distillation mechanism to propagate cumulative ordinal structure across network layers, improving hierarchical consistency of ordinal representations.
Extensive experiments on four general ordinal regression tasks demonstrate that \modelname consistently outperforms existing methods, particularly under noisy and imbalanced settings.
These results highlight the importance of moving beyond static supervision toward dynamic distribution refinement for robust ordinal representation learning. 
The code will be publicly available.

\end{abstract}




\maketitle


\section{Introduction}
\label{sec:intro}

\begin{figure}[t]
  \centering
  \includegraphics[width=0.92\linewidth]{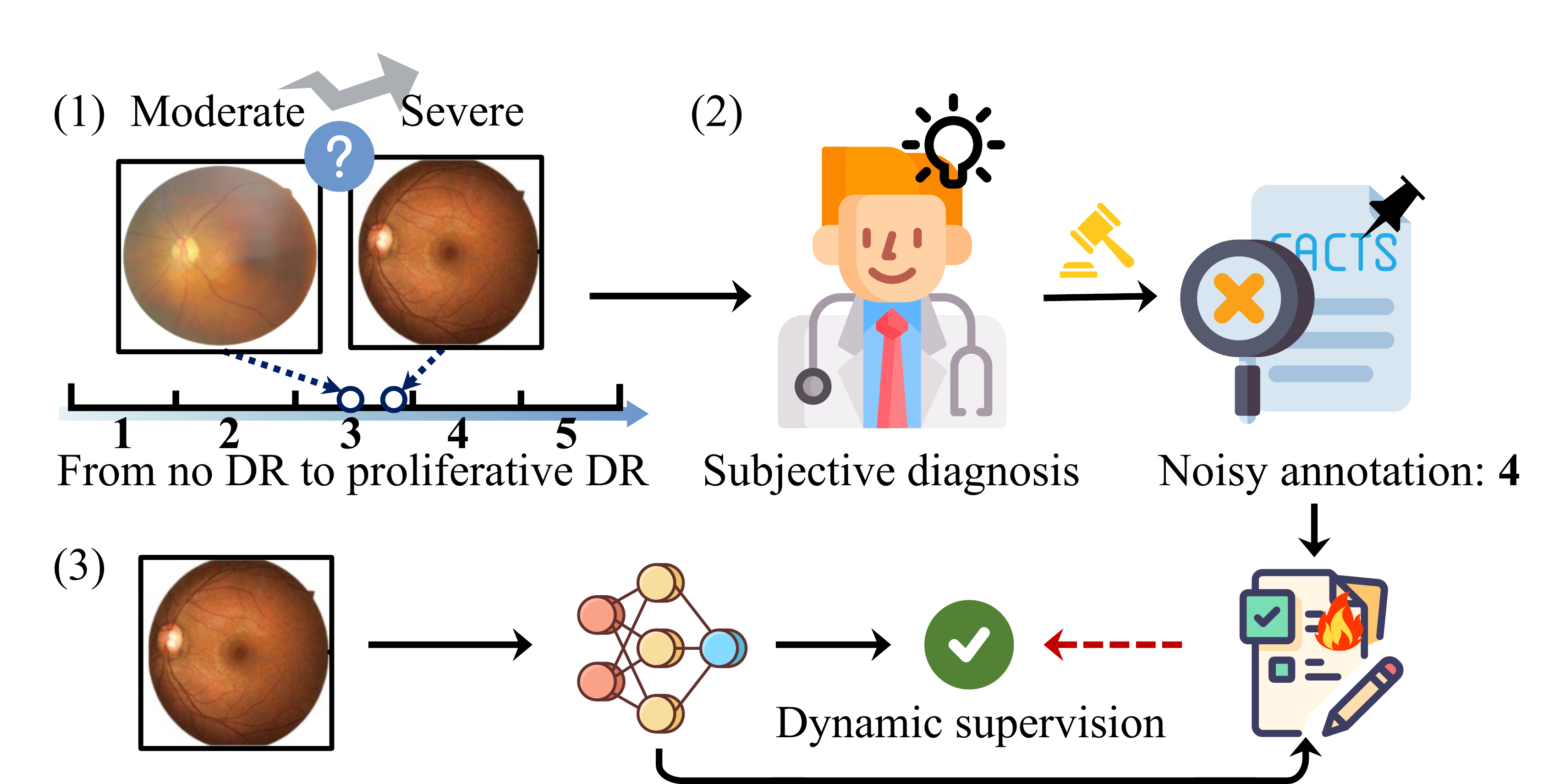}
  \caption{Motivation of dynamic supervision for ordinal regression. (1) Discretizing continuous attributes produces ambiguous class boundaries. (2) Subjective annotation introduces noisy labels. (3) Dynamic evolution of label distributions during training.
}
  \label{fig:introduction}
  \vspace{-3ex}
\end{figure}

Ordinal regression lies between regression and classification, focusing on prediction problems where labels are discrete but inherently ordered.
Such ordered labels are common in computer vision tasks.
For example, age is typically represented as integer-valued labels (\eg, 23, 24, 25), and disease severity is often categorized into progressive stages (\eg, mild, moderate, severe), in which the relative ordering among categories conveys essential information.
Consequently, image ordinal regression has become a fundamental learning problem in many interdisciplinary fields, including health (\eg, facial age estimation~\cite{Li_2019_CVPR,adience, cig}), bioinformatics (\eg, toxicity grades~\cite{Isaksson2020MLToxicity}), environmental (\eg, water quality levels~\cite{water}), and culture (\eg, image aesthetic assessment~\cite{10.1007/978-3-319-46448-0_40,Lee2019ImageAA}).
Compared to conventional classification, ordinal regression requires preserving inter-class ordering information, while differing from regression in that the target space remains categorical rather than continuous. 
As a result, the supervision signal in ordinal regression is inherently more structured, and effective learning critically depends on how ordinal label information is represented and utilized.


Existing ordinal regression methods primarily operate under a {static supervision}, 
where the target label is predefined and remains fixed throughout training. Depending on how the ordinal structure is derived from these static labels, these methods fall into two classifications. The first, the continuous space discretization (CSD) paradigm~\cite{wang2025survey}, assuming that ordinal labels arise from discretizing an underlying continuous variable, encodes ordinal structure through predefined loss functions~\cite{Pitawela_2025_CVPR} or label representations~\cite{chu2005gaussian, Li_2019_CVPR}.
The second encodes ordinality explicitly through structural reformulation, casting ordinal regression as ranking problems~\cite{Lim2020Order, Shin_2022_CVPR}, sequence prediction tasks~\cite{Ord2Seq_2023_ICCV, anonymous2026gor}, or distribution ordering problems~\cite{oldl, shaik2025ordinal}, thereby enforcing monotonic progression or step-wise decision processes across categories. 
Despite their differences, both paradigms share a fundamental limitation: since the label is static, any ordinal structure derived from it remains equally fixed and globally consistent, implicitly assuming that ordinal relations are uniformly reliable across all samples.


In practice, ordinal labels are obtained by discretizing an underlying continuous attribute through subjective human judgment, a process that inevitably compresses gradual semantic variation into a small number of discrete categories. Consequently, adjacent classes are inherently ambiguous and susceptible to annotation noise, which introduces systematic bias in both prediction and representation learning, hindering the recovery of the continuous ordinal structure~\cite{wang2025survey, Pitawela_2025_CVPR}.
For example, in diabetic retinopathy grading, adjacent severity levels are distinguished by subtle retinal changes, making borderline cases intrinsically confusable. This challenge is reflected in expert disagreement: in a screening study~\cite{jcm11113125} involving 670 eyes and 1501 retinal images, three retina specialists reached complete agreement on severity grading for only 58\% of eyes. 
A similar phenomenon arises in image aesthetic assessment, where ratings rely on subjective judgment influenced by personal taste, cultural background, and viewing context, resulting in inconsistent annotations even for the same image~\cite{Yang_2022_CVPR}.

These observations highlight that ordinal labels do not always provide perfectly reliable ordering information, and real-world data inevitably include samples with uncertain or noisy ordinal labels. 
However, most existing ordinal regression methods rely on static supervision schemes that treat labels or their derived distributions as fixed targets throughout training, and thus lack mechanisms to adapt to sample-dependent ambiguity and label noise as shown in Figure~\ref{fig:introduction}. Under subjective uncertainty, such fixed targets can force the model to adopt incorrect ordering constraints, with no way to revise them as learning progresses.
This can result in the model repeatedly reinforcing incorrect assumptions, potentially leading to suboptimal representations and degraded performance, especially in tasks where label uncertainty is prevalent.


To address these limitations, we propose \underline{D}ynamic \underline{D}istribution \underline{D}istillation for image \underline{o}rdinal regression (\modelname). 
In our approach, beyond relying solely on fixed ordinal targets, we incorporate an additional dynamic supervision that evolves with the model’s internal beliefs during training. 
Specifically, we leverage a self-distillation framework in which an auxiliary branch employs a contrastive ordinal-aware strategy to recover label distributions that capture both inter-class ambiguity and model confidence. 
These recovered distributions serve as dynamic soft supervision targets that are iteratively updated and used to guide the student model throughout training, progressively refining its understanding of the underlying ordinal structure.
Furthermore, to enhance the propagation of ordinal semantics across network layers, we propose a cross-layer interaction distillation strategy that enables shallow features to receive guidance from deeper ordinal representations, improving the alignment of ordinal features throughout the model hierarchy.
Our main contributions can be summarized as follows:

\begin{itemize}
\item We revisit ordinal regression from the perspective of supervision design and identify an inherent tension between globally fixed supervision and ambiguous ordinal annotations.
We argue that treating ordinal relations as deterministic and sample-invariant limits robustness under subjective noise and class imbalance, motivating a shift from static target fitting to adaptive supervision evolution.
\item We design a training-driven evolution of label distributions via self-distillation. 
Through contrastive ordinal-aware label enhancement, the model recovers refined distributions that capture instance-level ambiguity, while a CDF-based cross-layer interaction distillation mechanism propagates cumulative ordinal structure across network layers to ensure hierarchical consistency.
\item Extensive experiments and qualitative analyses across diverse ordinal datasets demonstrate consistent and significant improvements over state-of-the-art methods, particularly under noisy and imbalanced settings, validating the effectiveness of dynamic supervision for robust ordinal representation learning.
\end{itemize}


\section{Related Work}
\label{sec:related}

\subsection{Ordinal Regression}
Ordinal regression aims to predict discrete labels with a natural order while avoiding the overly strong assumption of equal inter-class distances.
One classic framework was K-rank~\cite{krank}, which trained $K$-$1$ classifiers and formulated ordinal regression as a set of ordered binary classification tasks focusing on distinguishing adjacent categories.
Furthermore, some methods~\cite{7780901, 8099569} extended convolutional neural networks to serve as subclassifiers.
Later studies further extended ordinal regression by explicitly exploiting rank structure during optimization.
Order learning~\cite{Lim2020Order} proposed a pairwise comparator to categorize the relationship between two instances and inferred the ordinal label of a query sample by enforcing consistency across comparisons with reference instances.
Subsequent studies further improved this paradigm by disentangling order-related and identity-related factors for more reliable reference comparison~\cite{lee2021deep}, or by extending pairwise comparison into continuous relative-rank regression within adaptive search windows~\cite{Shin_2022_CVPR}.
Instead of directly estimating an absolute rank, these methods inferred ordinal labels by comparing an input instance against reference samples with known ranks.
In addition, CLOC~\cite{Pitawela_2025_CVPR} minimized a multi-margin loss that provided learnable margins for flexible decision boundaries.
More recently, Ord2Seq~\cite{Ord2Seq_2023_ICCV} and GoR~\cite{anonymous2026gor} reframed ordinal regression as a sequence prediction or generation process to explicitly capture ordinal dependencies.
With the development of pre-trained VLMs, some recent studies~\cite{numCLIP, ordinalCLIP, l2r} exploited semantic priors in the textual embedding space to align ordinal structure with visual representations via prompt learning.

From a probabilistic perspective, some approaches assumed an underlying latent continuous variable and used thresholding to connect it with observed ordinal labels, such as Gaussian processes for ordinal regression~\cite{chu2005gaussian}.
Different from this, POEs~\cite{poe} represented each instance as a Gaussian distribution in the latent space and imposed an ordinal distribution constraint, thereby enabling uncertainty-aware ordinal prediction.
SORD~\cite{sord} and OLDL~\cite{oldl} instead adopted distribution-based probabilistic supervision, which replaced one-hot targets with ordinal-aware label distributions to better model neighborhood relationships and ambiguous decision boundaries.
Recent analyses also studied representation degeneracy in ordinal models, such as neural collapse phenomena in cumulative link models~\cite{collapse}.
Unlike previous studies, which imposed ordinal constraints through fixed label encodings or predefined target distributions, our method learns a dynamic teacher distribution via self-distillation, enabling progressively refined supervision that better captured ambiguous ordinal annotations.

\begin{figure*}[t]
  \centering
  \includegraphics[width=0.96\textwidth]{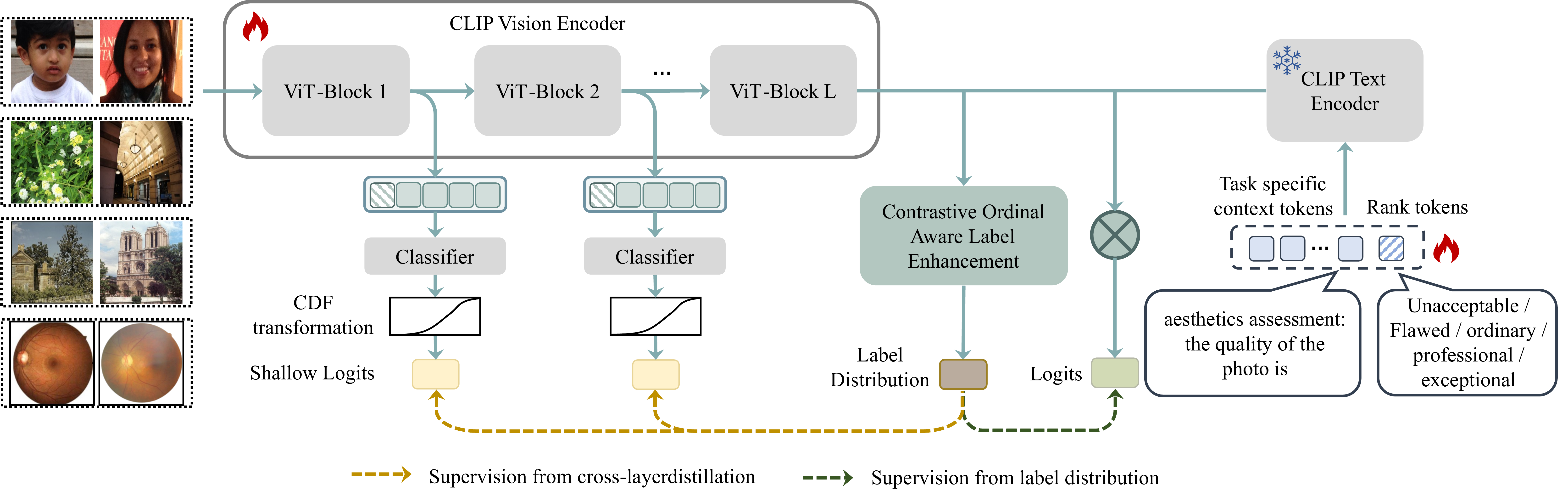}
  \caption{Overview of the proposed \modelname framework. 
Image and ordinal label embeddings are aligned in a shared vision–language space, where the contrastive ordinal-aware label enhancement module recovers instance-specific label distributions as the evolving teacher signals. 
The recovered distributions supervise the student prediction logits through self-distillation, while a CDF-based cross-layer interaction mechanism propagates cumulative ordinal structure to intermediate layers for consistent ordinal representation learning.
}
  \label{fig:framework}
\end{figure*}

\subsection{Label Distribution Learning}

Label distribution learning (LDL)~\cite{Geng_2016_TKDE} addresses supervision ambiguity by representing each instance with a distribution rather than a single hard label.
Early studies mainly established the general LDL paradigm and developed corresponding learning algorithms from the perspectives of problem transformation, algorithm adaptation, and specialized design \cite{Geng_2016_TKDE}.
For deep learning, DLDL extends LDL with distribution-based objectives to exploit label ambiguity and reduce overfitting on limited or noisy data~\cite{gao2017deep}.
LDL was instantiated in practical settings for apparent age estimation using deep age-distribution targets~\cite{Geng_2013_TPAMI,Yang_2015_ICCV_Workshops,Huo_2016_CVPR_Workshops}, and unified frameworks have been explored to jointly learn distributions and expected regress values~\cite{Gao_2018_IJCAI}. Since many LDL applications involve ordered labels, OLDL~\cite{oldl} explicitly studies ordinal label distribution learning  by modeling sequential patterns and order-consistent objectives. 

In addition, label enhancement (LE) studied a more practical setting where ground-truth label distributions were unavailable and needed to be recovered from logical labels. 
Early work formalized LE and proposed graph-based recovery methods to exploit feature topology and label correlation~\cite{Xu2021LE}. 
Later, variational label enhancement modeled latent label distributions through variational inference~\cite{Xu2020VLE}, while more recent methods further improved recovery quality by filtering label-irrelevant information via an information bottleneck objective~\cite{Zheng2023LIB}. 
In our paper, we follow label enhancement and focus on how to recover the label distribution that is both order-aware and ambiguity-aware.

\subsection{Self-distillation for Discriminative Tasks}
Knowledge distillation (KD) improves discriminative models by transferring ``dark knowledge'' via soft targets, often leading to better accuracy and calibration without increasing inference cost~\cite{hinton2015distilling}.
Beyond the standard teacher--student setting, {self-distillation} (self-KD) reuses the same model (or an EMA/previous-generation copy) as the teacher, and remains effective even when teacher and student share identical architectures~\cite{furlanello2018bornagain,yun2020regularizing}.

Recent studies increasingly position self-distillation as a {task-driven} tool for concrete discriminative applications rather than a generic regularizer. For example, MixSKD~\cite{mixskd} integrates Mixup into self-distillation to improve representation smoothness and transferability in generic visual classification.
In fine-grained recognition, where subtle inter-class differences and large intra-class variations co-exist, self-distillation is often instantiated as multi-level feature/logit supervision to enhance part-aware, multi-scale, and channel-selective representations, thereby strengthening discriminative features and reducing overfitting under limited data~\cite{hu2025hsd,demidov2024adnet}.
Similar design principles extend to other discriminative tasks (e.g., detection with noisy annotations), where self-distillation injects additional consistency constraints to stabilize training and improve robustness~\cite{Wu_2023_ICCV}. 
Unlike prior self-distillation methods that mainly regularize predictions or features, our method uses self-distillation to dynamically refine ambiguity-aware ordinal supervision and propagate cumulative ordinal structure across layers.

\section{Methodology}

\subsection{Problem Definition} 

Unlike standard multi-class classification, where class labels are unordered, ordinal regression requires respecting the intrinsic ordering among categories, such that the distance between labels carries semantic meaning.
Formally, given a dataset $\mathcal{D}=\{(x_i,y_i)\}_{i=1}^N$, where $N$ is the number of samples and $y_i$ denotes the ordinal rank of image $x_i$, the goal is to learn a shared embedding space that aligns image features with ordinal label embeddings for prediction via similarity.
For each input \(x_i\), we obtain the image embedding \(z_i = f_\theta(x_i)\), and predict the ordinal label by maximizing cosine similarity between \(z_i\) and each ordinal label embedding \(c_k\):
\begin{equation}
\hat{y}_i = \arg\max_{k\in\{1,\dots,K\}}\; \frac{z_i^\top c_k}{\|z_i\|\cdot \|c_k\|},
\end{equation}
where $K$ is the total number of ordinal classes. 
The predictive distribution over ordinal classes is then obtained by applying a softmax to the normalized cosine similarity logits:
\begin{equation}
p_\theta(y=k\mid x_i)
=
\frac{\exp\!\big( \frac{z_i^\top c_k}{\|z_i\|\cdot \|c_k\|}\big)}
{\sum_{j=1}^K \exp\!\big( \frac{z_i^\top c_j}{\|z_i\|\cdot \|c_j\|}\big)}.
\end{equation}

\subsection{Overall Self-Distillation Framework} \label{sec:distill}

In traditional ordinal regression, the supervision is directly derived from the ground-truth label (\eg, one-hot encoding), and the model is optimized by minimizing the discrepancy between the predictive distribution \(p_\theta(y\mid x)\) and this fixed supervision signal:
\begin{equation}
\mathcal{L}_{\text{static}}
=
\mathbb{E}_{(x_i,y_i)\sim\mathcal{D}}\;
\ell\big(p_\theta(y_i\mid x_i),\, y_i\big),
\end{equation}
where \(\ell(\cdot,\cdot)\) denotes an ordinal regression loss such as cross-entropy.  
This formulation implicitly assumes that the provided labels fully and reliably encode the underlying ordinal structure.
However, in many real-world scenarios, ordinal labels are discretized from continuous semantics and inevitably exhibit subjective ambiguity and annotation noise. 
Under such conditions, static supervision can impose incorrect ordering constraints on samples and repeatedly reinforce unreliable ordinal assumptions throughout training. 

To address this limitation, we introduce a self-distillation framework tailored for ordinal regression, as illustrated in Figure~\ref{fig:framework}. 
Instead of relying solely on fixed supervision, we construct an evolving teacher distribution \(q_t(y\mid x)\) from the model’s current representation and use it to supervise the student prediction \(p_\theta(y\mid x)\). 
The training objective becomes
\begin{equation}
\mathcal{L}_{\text{dynamic}}
=
\mathbb{E}_{(x_i,y_i)\sim\mathcal{D}}
\;
\ell\big(p_\theta(y_i\mid x_i),\, q_t(y_i\mid x_i)\big),
\end{equation}
where \(q_t(y\mid x)\) serves as a dynamically refined supervision signal. Note that the teacher model is updated through an exponential moving average (EMA) of the student parameters.
Unlike label-derived soft targets, the teacher distribution is dynamically derived from the model’s evolving representation during training. As a result, it captures instance-specific uncertainty and soft relations among neighboring ordinal categories that static supervision cannot express. Self-distillation thus enables adaptive supervision under noisy annotations, progressively correcting unreliable ordinal assumptions and improving alignment with latent ordinal semantics.


The proposed framework is implemented through two key modules. Section~\ref{COLE} presents the construction of the teacher distribution \(q_t(y\mid x)\) for ordinal-aware dynamic supervision, while Section~\ref{cdfd} introduces a CDF-based cross-layer interaction distillation mechanism that propagates ordinal structure across network layers.

\subsection{Contrastive Ordinal Aware Label Enhancement} \label{COLE}

\begin{figure}[t]
  \centering
  \includegraphics[width=0.96\linewidth]{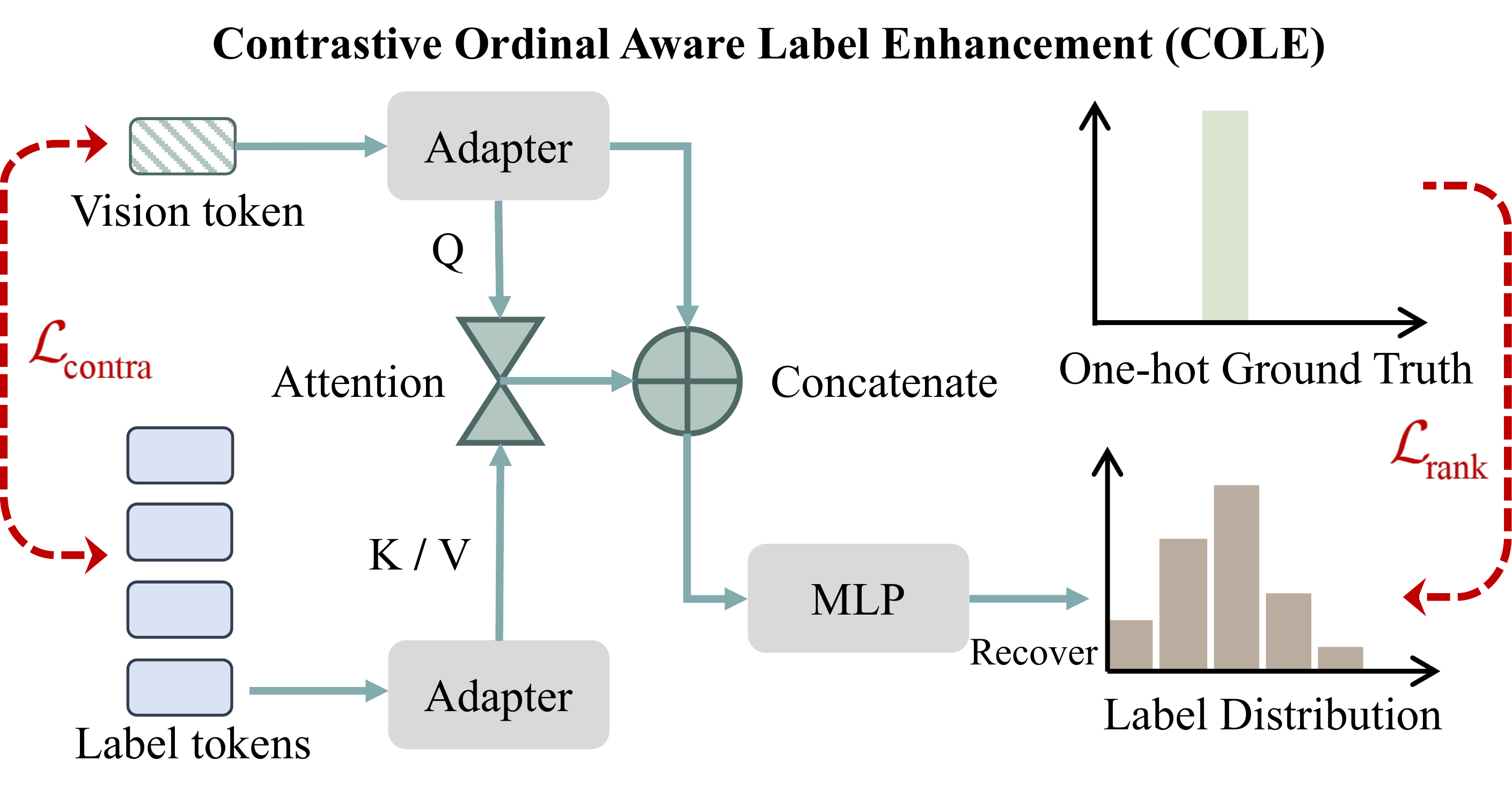}
  \caption{Illustrating the Contrastive Ordinal Aware Label Enhancement module.}
  \label{fig:cole}
\end{figure}

In this subsection, we describe the construction of the teacher distribution \(q_t(y\mid x)\). As illustrated in Figure~\ref{fig:cole}, COLE recovers an instance-specific ordinal distribution that captures ordinal semantics, 
thereby serving as the dynamic supervision signal in the self-distillation framework.

\textbf{Contrastive Ordinal-Structured Representation Learning.}
Existing evidence suggests that text embeddings of ordinal labels already exhibit meaningful relative ordering in the semantic space, making them a natural carrier of ordinal information~\cite{ordinalCLIP, numCLIP}. Motivated by this observation, we adopt a contrastive learning paradigm to align image embeddings with their corresponding ordinal label embeddings.
Given an input image \(x_i\), its visual embedding is computed as \(z_i = f_\theta(x_i)\in\mathbb{R}^d\), and each ordinal class is associated with a text embedding \(c_k\in\mathbb{R}^d\). 
The contrastive objective encourages the embedding of \(x_i\) to be close to its ground-truth ordinal label embedding \(c_{y_i}\) while being separated from other label embeddings:
\begin{equation}
\mathcal{L}_{\text{contra}}
=
\mathbb{E}_{(x_i,y_i)\sim\mathcal{D}}
\left[
-\log
\frac{\exp\!\big(\mathrm{sim}(z_i, c_{y_i})/\tau\big)}
{\sum_{k=1}^K\exp\!\big(\mathrm{sim}(z_i, c_{k})/\tau\big)}
\right],
\end{equation}
where \(\mathrm{sim}(z,t)\) denotes cosine similarity and \(\tau>0\) is a temperature hyperparameter. 
This alignment induces a structured representation space in which ordinal relationships are encoded in the relative geometry of label embeddings~\cite{numCLIP}.

\textbf{Attention-Based Distribution Recovery.}
To recover instance-specific ordinal distributions from the structured representation space, we explicitly model the interaction between the image feature $z_i$ and the ordinal label embeddings ${c_k}$. Specifically, we use a scaled dot-product attention mechanism, where the image embedding serves as the query and the ordinal label embeddings serve as the keys and values.
The attention weights are computed as:
\begin{equation}
\alpha_{i,k} 
=
\frac{\exp\!\big((W_q z_i)^\top (W_k c_k) / \sqrt{d}\big)}
{\sum_{j=1}^K \exp\!\big((W_q z_i)^\top (W_k c_j) / \sqrt{d}\big)},
\end{equation}
where \(W_q\) and \(W_k\) are learnable projection matrices. 
The class-aware aggregated token is then obtained by
\begin{equation}
\tilde{c}_i
=
\sum_{k=1}^K \alpha_{i,k} \, c_k .
\end{equation}

We then concatenate the aggregated token with the image embedding and feed the joint representation into a multi-layer perceptron to obtain the recovered logits:
\begin{equation}
s_i = \mathrm{MLP}([z_i \Vert \tilde{c}_i]).
\end{equation}
The teacher distribution is then defined as:
\begin{equation}
q_t(y\mid x_i)
=
\mathrm{Softmax}(s_i),
\end{equation}
where \(q_t(y\mid x_i)\) integrates image-specific evidence with the ordinal semantics encoded in the label embeddings, thereby serving as the dynamic supervision signal in the self-distillation framework.

\textbf{Ordinal-Regularized Distribution Learning.}
While the recovered logits \(s_i\) are optimized through contrastive learning, 
the resulting distribution may still violate ordinal structure, \eg, assigning higher confidence to categories that are far from the true rank. 
To explicitly enforce ordinal consistency, we introduce a distance-weighted ranking constraint on the recovered logits.

Formally, let \(y_i\) denote the ground-truth ordinal index of sample \(x_i\). 
For each non-ground-truth category \(k \neq y_i\), we penalize violations in which its logit exceeds that of the true category, 
with a penalty proportional to the ordinal distance \(|k - y_i|\):

\begin{equation}
\mathcal{L}_{\text{rank}}
=
\mathbb{E}_{(x_i,y_i)\sim\mathcal{D}}
\sum_{k \neq y_i}
|k - y_i|
\max\big(0,\; s_{i,k} - s_{i,y_i}\big).
\end{equation}

The distance weighting term  \(|k-y_i|\) encodes an ordinal prior, such that mis-rankings involving distant categories incur larger penalties than those between adjacent ranks.
This formulation explicitly injects ordinal geometry into the distribution learning process, encouraging unimodal behavior centered on the true rank while allowing mild uncertainty among neighboring categories.




\subsection{CDF-Based Cross-Layer Interaction Distillation} \label{cdfd}

While the evolving distribution \(q_t(y\mid x)\) provides dynamic supervision at the final prediction layer, 
ordinal semantics may emerge at different layers of the visual encoder.
Relying only on final-layer supervision can therefore provides insufficient ordinal guidance to shallow layers, potentially limiting the alignment of shallow representations with the underlying ordinal structure.
To address this issue, we introduce a cross-layer interaction distillation mechanism that transfers ordinal information from the dynamically recovered teacher distribution \(q_t(y\mid x)\) to multiple intermediate layers.

Since ordinal labels arise from discretizing an underlying continuous variable, their intrinsic structure is more faithfully captured by cumulative relations rather than independent class probabilities.
In particular, ordinal consistency is naturally encoded by cumulative probabilities over ordinal thresholds, which preserve the monotonic ordering among ranks.
Motivated by this property, we adopt a cumulative distribution function (CDF) formulation to encode ordinal structure in cross-layer distillation.

Formally, let \(p^{(l)}_\theta(y\mid x)\) denote the predictive distribution from the \(l\)-th layer after applying a classification head and softmax function. 
Rather than directly aligning class probabilities across layers, we explicitly transfer ordinal structure via cumulative representations.
For each layer \(l\), we convert its predictive distribution into a cumulative distribution function (CDF) form:
\begin{gather}
\mathbf{c}^{(l)}(x) = \big[ P^{(l)}(y \le 1 \mid x), \dots, 
P^{(l)}(y \le K \mid x)
\big], \nonumber
\\ 
P^{(l)}(y \le k \mid x) = \sum_{j=1}^{k} p^{(l)}_\theta(y=j\mid x).
\end{gather}
In the same manner, we transform the teacher distribution into its cumulative form:
\begin{gather}
\mathbf{c}^{(q)}(x)
=
\big[
P^{(q)}(y \le 1 \mid x),
\dots,
P^{(q)}(y \le K \mid x)
\big], \nonumber
\\
P^{(q)}(y \le k \mid x)
=
\sum_{j=1}^{k}
q_t(y=j\mid x).
\end{gather}
We then enforce cross-layer cumulative consistency as follows:
\begin{equation}
\mathcal{L}_{\text{cdf-dist}} = \mathbb{E}_{(x_i,y_i)\sim\mathcal{D}}\sum_{l \in \mathcal{S}} \left\|
\mathbf{c}^{(l)}(x_i) - \text{stopgrad}\!\left(\mathbf{c}^{(q)}(x_i)\right) \right\|_2^2,
\end{equation}
where \(\mathcal{S}\) denotes the set of supervised intermediate layers and 
\(\text{stopgrad}(\cdot)\) blocks gradient flow to the teacher branch.


Unlike conventional deep--shallow distillation, which transfers logits or class probabilities directly, our formulation distills cumulative ordinal structure in the CDF space. This encourages intermediate layers to preserve global rank consistency instead of merely matching isolated class scores. As a result, shallow representations inherit structurally coherent ordinal information from the dynamically refined teacher distribution\(q_t(y\mid x)\), thereby improving representation alignment and enhancing robustness under ambiguous and noisy supervision.



\subsection{Overall Objective}

Our training objective consists of three complementary terms:
(1) Static Ordinal Supervision,
(2) Contrastive Ordinal Aware Label Enhancement,
and (3) Self-Distillation with Cross-Layer Interaction.

{(1) Static Ordinal Supervision.}
To preserve consistency with ground-truth annotations, 
we retain the cross-entropy loss:
\begin{equation}
\mathcal{L}_{\mathrm{CE}}
=
\mathbb{E}_{(x_i,y_i)\sim\mathcal{D}}
\;
\ell_{\text{CE}}\big(p_\theta(y_i\mid x_i),\, y_i\big),
\end{equation}
where $p_\theta(y\mid x)$ denotes the final-layer prediction.
This term provides a stable supervision anchor and prevents the dynamic teacher from drifting during early training.

{(2) Contrastive Ordinal-Aware Label Enhancement.}
To recover refined ordinal distributions, we employ the COLE objective:
\begin{equation}
\mathcal{L}_{\text{COLE}}
=
\alpha_1 \mathcal{L}_{\text{contra}}
+
\alpha_2 \mathcal{L}_{\text{rank}},
\end{equation}
which aligns image and label embeddings while enforcing ordinal consistency in the recovered distribution.

{(3) Self-Distillation with Cross-Layer Interaction.}
Using the dynamically recovered distribution $q_t(y\mid x)$ as the teacher signal,
we perform self-distillation at the final prediction layer together with CDF-based cross-layer interaction:
\begin{equation}
\mathcal{L}_{\text{SD}}
=
\lambda_1 \mathbb{E}_{(x_i,y_i)\sim\mathcal{D}}
\;
\ell_{\mathrm{KL}}\!\left(
q_t(y_i\mid x_i)
\,\Vert\,
p_\theta(y_i\mid x_i)
\right)
+
\lambda_2
\mathcal{L}_{\text{cdf-dist}},
\end{equation}
where $\ell_{\mathrm{KL}}$ denotes the KL divergence, and $\mathcal{L}_{\text{cdf-dist}}$ enforces cumulative ordinal consistency across intermediate layers.

Accordingly, the full training objective is
\begin{equation}
\mathcal{L}
=
\mathcal{L}_{\text{CE}}
+
\mathcal{L}_{\text{COLE}}
+
\mathcal{L}_{\text{SD}}.
\end{equation}

\section{Experiments}

We conduct extensive experiments to validate the effectiveness and robustness of the proposed \modelname\ across four representative ordinal regression tasks: Image Aesthetics Assessment, Facial Age Estimation, Historical Image Dating, and Diabetic Retinopathy Grading.

\subsection{Experimental Setup}

We implement our framework in PyTorch and conduct all experiments using a server equipped with NVIDIA RTX 4090 GPUs.
We adopt CLIP ViT-B/16 as the backbone model for all experiments, including both the visual encoder and the corresponding text encoder.
Following previous work~\cite{numCLIP}, all input images are first resized to $256 \times 256$ and then randomly cropped to $224 \times 224$ during training. 
The teacher model is updated via an exponential moving average (EMA) of the student parameters with momentum $m$.
The student model is optimized using the Adam optimizer with a learning rate of $1\times10^{-4}$ and a batch size of 64.

\begin{table*}[t]
\centering
\setlength{\tabcolsep}{8pt}
\renewcommand{\arraystretch}{1.1}
\caption{Results on Image Aesthetics dataset}
\label{tab:aesthetics}
\begin{tabular}{l ccccc ccccc}
\toprule
\multirow{2}{*}{Methods} &
\multicolumn{5}{c}{Accuracy(\%, $\uparrow$)} &
\multicolumn{5}{c}{MAE($\downarrow$)} \\
\cmidrule(lr){2-6}\cmidrule(lr){7-11}
& Nature & Animal & Urban & People & Overall
& Nature & Animal & Urban & People & Overall \\
\midrule
CNNPOR~\cite{cnnpor}       & 71.86 & 69.32 & 69.09 & 69.94 & 70.05 & 0.294 & 0.322 & 0.325 & 0.321 & 0.316 \\
SORD~\cite{sord}           & 73.59 & 70.29 & 73.25 & 70.59 & 72.03 & 0.271 & 0.308 & 0.276 & 0.309 & 0.290 \\
POE~\cite{poe}          & 73.62 & 71.14 & 72.78 & 72.22 & 72.44 & 0.273 & 0.299 & 0.281 & 0.293 & 0.287 \\
GOL~\cite{NEURIPS2022_00358de3}          & 73.8  & 72.4  & 74.2  & 69.6  & 72.7  & 0.27  & 0.28  & 0.26  & 0.31  & 0.28  \\
\midrule
Zero-shot CLIP~\cite{pmlr-v139-radford21a}  & 65.24 & 45.67 & 58.78 & 53.06 & 55.68 & 0.461 & 0.557 & 0.468 & 0.524 & 0.502 \\
CoOp~\cite{10.1007/s11263-022-01653-1}          & 72.74 & 71.46 & 72.14 & 69.34 & 71.42 & 0.285 & 0.298 & 0.294 & 0.330 & 0.302 \\
OrdinalCLIP~\cite{ordinalCLIP}  & 73.65 & 72.85 & 73.20 & 72.50 & 73.05 & 0.273 & 0.279 & 0.277 & 0.291 & 0.280 \\
L2RCLIP~\cite{l2r}    & 73.51 & \underline{75.26} & 77.76 & \underline{78.69} & 76.07 & 0.267 & 0.253 & 0.216 & \underline{0.246} & 0.245 \\
NumCLIP~\cite{numCLIP} & \underline{75.20} & 75.24& \underline{79.49} & 76.17 & \underline{76.53}& \underline{0.249} & \underline{0.250} & \underline{0.208} & \textbf{0.238} & \underline{0.236} \\
\rowcolor{lightpurple}\modelname (Ours) & \textbf{77.05} & \textbf{76.51} & \textbf{80.08} & \textbf{76.51} & \textbf{77.54} & \textbf{0.229} & \textbf{0.238} & \textbf{0.202} & \textbf{0.238} & \textbf{0.227} \\ 
\bottomrule
\end{tabular}

\end{table*}

{\captionsetup{singlelinecheck=false, justification=centering}

\begin{table*}[t]
\centering
\begin{minipage}[t]{0.32\textwidth}
\centering
\renewcommand{\arraystretch}{1.1}
\caption{Results on MORPH II dataset. }
\label{tab:morph}
    \begin{tabular}{l c}
    \toprule
    \multirow{2}{*}{Methods} & \multicolumn{1}{c}{MOROH II} \\
    \cmidrule(lr){2-2}
    & MAE ($\downarrow$) \\
    \midrule
    AGEn~\cite{8141981}         & 2.52 \\
    BridgeNet~\cite{Li_2019_CVPR}     & 2.38 \\
    AVDL~\cite{10.1007/978-3-030-58592-1_23}          & 2.37 \\
    POE~\cite{poe}           & 2.35 \\
    PML~\cite{Deng_2021_CVPR}            & 2.15 \\
    MWR~\cite{Shin_2022_CVPR}        & 2.13 \\
    \midrule
    Zero-shot CLIP~\cite{pmlr-v139-radford21a}  & 6.91  \\
    CoOp~\cite{10.1007/s11263-022-01653-1}          & 2.39  \\
    OrdinalCLIP~\cite{ordinalCLIP} & 2.32 \\
    L2RCLIP~\cite{l2r}      & 2.13 \\
    NumCLIP~\cite{numCLIP} & \underline{2.08} \\
    \rowcolor{lightpurple}\modelname (Ours) & \textbf{2.03} \\
    \bottomrule
    \end{tabular}
\end{minipage}
\hfill
\begin{minipage}[t]{0.66\textwidth}
\centering
\renewcommand{\arraystretch}{1.1}
\caption{Results on Adience and HCI dataset.}
\label{tab:adience}
    \begin{tabular}{l cc cc}
    \toprule
    \multirow{2}{*}{Methods} & \multicolumn{2}{c}{Adience}& \multicolumn{2}{c}{HCI} \\
    \cmidrule(lr){2-3} \cmidrule(lr){4-5}
    & Accuracy (\%, $\uparrow$) & MAE ($\downarrow$)& Accuracy (\%, $\uparrow$) & MAE ($\downarrow$) \\
    \midrule
    CNNPOR~\cite{cnnpor}         & $57.4 \pm 5.8$  & $0.55 \pm 0.08$& $50.1\pm 2.7$  & $0.82 \pm 0.05$ \\
    SORD~\cite{sord}            & $59.6 \pm 3.6$  & $0.49 \pm 0.05$& $53.4 \pm 3.7$  & $0.70 \pm 0.05$ \\
    POE~\cite{poe}            & $60.5 \pm 4.4$  & $0.47 \pm 0.06$& $54.7 \pm 3.2$  & $0.66 \pm 0.05$ \\
    MWR~\cite{Shin_2022_CVPR}               & 62.6           & 0.45& $57.8 \pm 4.1$  & $0.58 \pm 0.05$ \\
    Ord2Seq~\cite{Ord2Seq_2023_ICCV} & 63.9 & 0.43& $60.9 \pm 1.6$ & $ 0.52 \pm 0.01$ \\
    CLOC~\cite{Pitawela_2025_CVPR} & 63.0 & 0.41& 62.1 & 0.55 \\
    \midrule
    Zero-shot CLIP~\cite{pmlr-v139-radford21a} & $43.3 \pm 3.6$ & $0.80 \pm 0.02$ & $30.4 \pm 3.3$ & $1.01 \pm 0.03$\\
    CoOp~\cite{10.1007/s11263-022-01653-1}         & $60.6 \pm 5.5$ & $0.50 \pm 0.08$& $51.9 \pm 2.6$ & $0.76 \pm 0.06$ \\
    OrdinalCLIP~\cite{ordinalCLIP} & $61.2 \pm 4.2$ & $0.47 \pm 0.06$ & $56.4 \pm 1.7$ & $0.67 \pm 0.03$\\
    L2RCLIP~\cite{l2r}    & $\underline{66.2 \pm 4.4}$ & ${0.36 \pm 0.05}$ & ${67.2 \pm 1.6}$ & ${0.43 \pm 0.02}$ \\
    NumCLIP~\cite{numCLIP} & $65.7 \pm 4.6$ &  $ \underline{0.34 \pm 0.06}$& $\underline{69.6 \pm 2.0}$ & $\underline{0.35 \pm 0.03}$ \\
    \rowcolor{lightpurple}\modelname (Ours) & $\mathbf{69.7 \pm 4.8}$& $\mathbf{0.32 \pm 0.05}$ & {$\mathbf{71.7 \pm 1.2}$} & {$\mathbf{0.32 \pm 0.01}$}\\
    \bottomrule
    \end{tabular}
\end{minipage}
\end{table*}

}


\subsection{Overall Performance}

\subsubsection{Image Aesthetics Assessment}

The Image Aesthetics dataset~\cite{Schifanella_Redi_Aiello_2021} contains 13,706 publicly available Flickr images spanning four semantic categories: {animals}, {urban}, {people}, and {nature}. 
Each image is annotated by at least five human graders with a discrete aesthetic score ranging from 1 to 5, corresponding to five ordered levels of photographic quality.
The ground-truth label is defined as the median of all assigned ratings to reduce annotation variance.
Following~\cite{numCLIP, fuzzy-ord}, we adopt 5-fold cross-validation for fair comparison with prior methods. 
Both mean absolute error (MAE) and classification accuracy are reported for evaluation.

\textbf{Performance.}
Table~\ref{tab:aesthetics} summarizes the quantitative comparison with state-of-the-art methods on the Image Aesthetics dataset.
Our method achieves the best overall performance, obtaining the lowest MAE of 0.227 and the highest overall accuracy of 77.5\%. 
These gains are particularly meaningful for image aesthetics assessment, where labels are aggregated from multiple human judgments and thus inherently subject to ambiguity. Because adjacent quality levels often share highly overlapping visual characteristics, boundary cases are difficult to separate under fixed supervision.
It also consistently improves results across all four semantic categories, with the largest gain observed in the Nature category, where accuracy increases from 75.20\% to 77.05\%. These results indicate that the recovered ordinal distribution helps the model better capture subtle aesthetic cues in visually diverse scenes.
Compared with recent CLIP-based approaches, our method demonstrates additional gains, highlighting the effectiveness of the proposed dynamic supervision and ordinal-aware enhancement mechanisms beyond backbone pretraining.
The consistent improvements in both MAE and accuracy further suggest that our method not only increases exact prediction accuracy, but also reduces the severity of ordinal errors by modeling ambiguous samples more effectively.

\subsubsection{Facial Age Estimation}

We evaluate our method on two widely used facial age estimation datasets: MORPH II~\cite{morph} and Adience~\cite{adience}.
MORPH II consists of 55,134 face images with ages spanning from 16 to 77 years, and is widely used for fine-grained age estimation.
Adience contains 26,580 facial images annotated with eight discrete age groups.
Following~\cite{ordinalCLIP}, we adopt the standard evaluation protocol and the commonly used train/test split for both datasets. Mean absolute error (MAE) is reported on both datasets, and classification accuracy is additionally reported on Adience.

\textbf{Performance.}
Tables~\ref{tab:morph} and~\ref{tab:adience} summarize the comparison results on the MORPH II and Adience, respectively. 
Our method achieves state-of-the-art performance across both datasets.
On MORPH II, \modelname reduces MAE from 2.08 to 2.03 compared with the strongest prior method, indicating more accurate fine-grained age estimation.
On Adience, it improves classification accuracy from 66.2\% to 69.7\%, while also reducing MAE from 0.34 to 0.32, demonstrating better alignment with coarse-grained age groups.
Moreover, although CLIP-based methods already benefit from large-scale pretraining, our method still achieves clear improvements over recent CLIP-based ordinal regression baselines, indicating that the gain comes not only from the backbone prior, but also from the proposed dynamic supervision and ordinal-aware enhancement mechanisms.

\begin{table}[t]
    \renewcommand{\arraystretch}{1.1}
    \centering
    \caption{Results on DR dataset.}
    \label{tab:dr}
    \begin{tabular}{l cc}
        \toprule
        \multirow{2}{*}{Methods} & \multicolumn{2}{c}{DR} \\
        \cmidrule(lr){2-3}
        & Accuracy (\%, $\uparrow$) & MAE ($\downarrow$) \\
        \midrule
        Poisson~\cite{poission}        & $77.1 \pm 0.6$  & $0.38 \pm 0.25$ \\
        MT~\cite{DBLP:journals/corr/abs-1805-11837}            & $82.8 \pm 0.6$  & $0.36 \pm 0.22$ \\
        SORD~\cite{sord}            & $78.2 \pm 0.6$  & $0.73 \pm 0.17$ \\
        POE~\cite{poe}            & $80.5 \pm 0.6$   & $0.30 \pm 0.21$ \\
        Ord2Seq~\cite{Ord2Seq_2023_ICCV} & $83.1 \pm 0.5$ & $0.25 \pm 0.07$ \\
        DFPG~\cite{fuzzy-ord} &  $83.4 \pm 0.5$& $0.24 \pm 0.13$ \\
        \midrule
        Zero-shot CLIP~\cite{pmlr-v139-radford21a} & $73.5 \pm 0.0$ & $0.52 \pm 0.00$ \\
        NumCLIP~\cite{numCLIP}    & $78.9 \pm 0.6$ & $0.38 \pm 0.21$\\
        \rowcolor{lightpurple}\modelname (Ours) & {$\mathbf{83.9 \pm 0.6}$} & {$\mathbf{0.23 \pm 0.10}$} \\
        \bottomrule
    \end{tabular}
\end{table}

\subsubsection{Historical Image Dating}
Historical Color Image~\cite{hci_2} is a dataset designed for decade-level prediction of historical color photos. 
The dataset spans five decades, from the 1930s to the 1970s, with 265 images in each decade.
Following previous work~\cite{numCLIP, Pitawela_2025_CVPR}, we adopt the standard ordinal regression setting and report both accuracy and MAE, together with their standard deviations.

\textbf{Performance.} 
The results on HCI are summarized in Table~\ref{tab:adience}.
Our method achieves the best performance among all compared approaches, reaching an accuracy of 71.7\% and an MAE of 0.32. Compared with the strongest prior method, it further achieves a 2.1\% accuracy improvement.
For CLIP-based baselines, although strong large-scale pretraining provides a powerful visual prior, their embedding space is not inherently aligned with the ordinal temporal semantics required for decade prediction in this dataset, leading to weak zero-shot performance.
CoOp~\cite{10.1007/s11263-022-01653-1} improves over zero-shot CLIP by learning task-specific prompts, but the gain remains limited, likely because the relatively small dataset makes it difficult for prompt tuning alone to establish robust ordinal alignment in the feature space.
In contrast, L2RCLIP and NumCLIP~\cite{l2r, numCLIP} achieve substantially better results by explicitly incorporating ordinal label relationships into representation learning.
Building on this line, our method further improves performance by introducing dynamic supervision, which better captures gradual temporal transitions and reduces confusion between neighboring decades.
 
\subsubsection{Diabetic Retinopathy Grading}
To further assess the robustness and generalization ability of our method in the presence of subjective ambiguity and annotation noise, we evaluate \modelname on the Diabetic Retinopathy (DR) dataset, which contains 35,126 high-resolution retinal fundus images annotated into five ordinal severity levels ranging from no DR to proliferative DR. 
Following the standard evaluation protocol~\cite{cig, fuzzy-ord}, we report both accuracy and MAE to evaluate classification performance and ordinal consistency.

\textbf{Performance.} Table~\ref{tab:dr} summarizes the results on the DR dataset. 
Our method achieves the best overall performance, reaching the highest accuracy of 83.9\% and the lowest MAE of 0.23 among all compared approaches.
Given the severe class imbalance in DR grading, 73.5\% of samples belong to grade 1 (no DR), which can disproportionately bias supervision toward low-severity representations and make reliable rank transitions among minority classes difficult to learn.
Especially, SORD~\cite{sord} constructs fixed distance-aware soft labels based solely from the ground-truth rank and yields relatively higher MAE of 0.73. 
These results suggest that dynamic supervision is particularly valuable when static label-derived constraints are dominated by majority-class patterns.

A different phenomenon is observed for CLIP-based baselines. Although CLIP-based methods perform strongly in tasks such as image aesthetics assessment and age estimation, they are markedly less effective on DR grading.
In particular, the reported accuracy of zero-shot CLIP in Table~\ref{tab:dr} is misleading: a closer inspection shows that it collapses to a trivial strategy that predicts every sample as grade 1 across all folds, so the resulting 73.5\% accuracy merely reflects the majority-class ratio rather than meaningful diagnostic capability.
This performance gap is likely due to the domain mismatch between CLIP pretraining and retinal imaging, since fundus images belong to a specialized medical domain with visual characteristics that are substantially underrepresented in natural image–text pairs.
In contrast, \modelname remains effective under this challenging setting. By introducing dynamic supervision, our method better models minority-grade transitions and avoids over-reliance on dominant low-severity labels. The consistent improvement over both conventional ordinal methods and CLIP-based baselines indicates that dynamic supervision is particularly beneficial for robust ordinal learning in highly skewed medical datasets.

\subsection{Further Analysis}

\subsubsection{Ablation Study}
\begin{table}[t]
\centering
\renewcommand{\arraystretch}{1.1}
\caption{Ablation study on MORPH II and Adience datasets.}
\label{tab:ablation}
\begin{tabular}{lcc c cc}
\toprule
\multicolumn{3}{c}{Setting}
& \multicolumn{1}{c}{MORPH II}
& \multicolumn{2}{c}{Adience} \\
\cmidrule(lr){1-3} \cmidrule(lr){4-4} \cmidrule(lr){5-6}
SD & COLE & CDF & MAE $\downarrow$
& Acc $\uparrow$ & MAE $\downarrow$ \\
\midrule
-&-&- & 2.35 & 64.1 & 0.42 \\
\checkmark &-&- & 2.18 & 66.0 & 0.36 \\
\checkmark&\checkmark&- & 2.08 & 67.6 & 0.35 \\
\rowcolor{lightpurple}\checkmark&\checkmark&\checkmark & \textbf{2.03} & \textbf{69.7} & \textbf{0.32} \\
\bottomrule
\end{tabular}
\end{table}

We conduct ablation experiments to evaluate the contribution of each proposed module in \modelname.
Table~\ref{tab:ablation} reports the results on MORPH II and Adience datasets by progressively enabling the proposed modules.

Starting from a baseline model built upon a CLIP backbone and trained with standard cross-entropy loss, introducing the self-distillation (SD) module already yields noticeable improvements, reducing the MAE on MORPH II from 2.35 to 2.18 and improving the accuracy on Adience from 64.1\% to 66.0\%.
This result highlights the benefit of dynamic supervision, where the evolving teacher distribution provides progressively refined soft targets during training.
Compared with static one-hot supervision, such dynamic supervision better captures ordinal uncertainty and stabilizes representation learning.
Adding the contrastive ordinal-aware label enhancement (COLE) further improves performance on both datasets. This suggests that explicitly modeling ordinal relationships in the shared vision--language space yields more informative supervision signals than relying on prediction-level supervision alone.
Finally, incorporating the proposed CDF-based supervision yields the best performance, achieving an MAE of 2.03 on MORPH II and an accuracy of 69.7\% on Adience.
The additional gain indicates that transferring cumulative ordinal structure to intermediate layers helps preserve intrinsic rank relationships throughout the network hierarchy.
Overall, the consistent gains obtained by progressively enabling SD, COLE, and CDF demonstrate that the three modules provide complementary benefits for robust ordinal learning.

\subsubsection{Robustness to Annotation Noise}

\begin{figure}[t]
  \centering
  \includegraphics[width=0.98\linewidth]{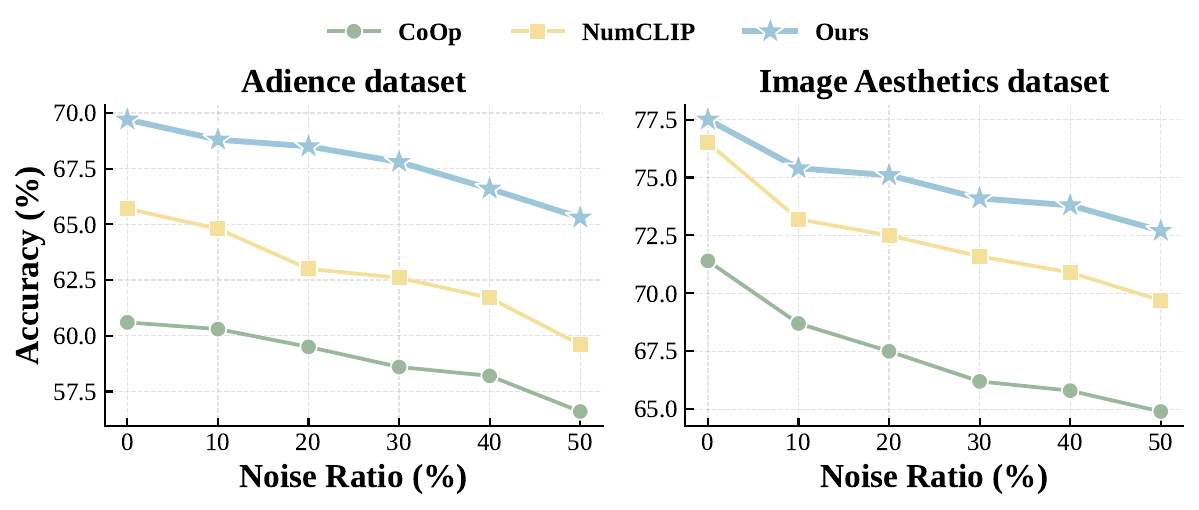}
  \caption{
    Robustness to label noise on the Adience and Image Aesthetics dataset.
    Training labels are perturbed with adjacent noise (±1) at ratios of
    10\%, 20\%, 30\%, 40\%, and 50\%.
    }
  \label{fig:noise}
\vspace{-1ex}
\end{figure}

To further evaluate the robustness of different methods under noisy supervision, we conduct an additional experiment on the Adience and Image Aesthetics dataset with artificially perturbed labels. 
Specifically, we introduce controlled label perturbations to the training set by randomly shifting a portion of labels to adjacent categories (±1), which simulates realistic annotation errors commonly observed in ordinal tasks.
The perturbation ratios are set to 10\%, 20\%, 30\%, 40\%, and 50\%, while the test set remains unchanged to ensure fair evaluation. 

Using these perturbed datasets, we train three representative methods, including CoOp~\cite{10.1007/s11263-022-01653-1}, NumCLIP~\cite{10.1007/s11263-022-01653-1, numCLIP}, and our proposed \modelname, and evaluate their performance under increasing noise levels.
As illustrated in Figure~\ref{fig:noise}, the performance of all methods decreases as the noise ratio increases on the both adience and image aesthetics dataset. 
However, our method degrades significantly slower than CoOp and NumCLIP, demonstrating stronger robustness to noisy supervision. 
This improvement stems from the dynamic supervision mechanism, where progressively refined teacher distributions provide more informative soft targets during training. 
Such evolving supervision reduces the impact of incorrect annotations and enables the model to learn more stable ordinal representations.

Overall, these results highlight the advantage of dynamic supervision for robust ordinal learning. 
Unlike static supervision schemes that rely directly on potentially corrupted labels, the evolving supervision signals in \modelname allow the model to maintain stable ordinal predictions even under severe label perturbations.
This observation further suggests that adaptive supervision is crucial for learning reliable ordinal structures in the presence of annotation noise.


\subsubsection{Visualization}

\begin{figure}[t]
  \centering
  \includegraphics[width=0.99\linewidth]{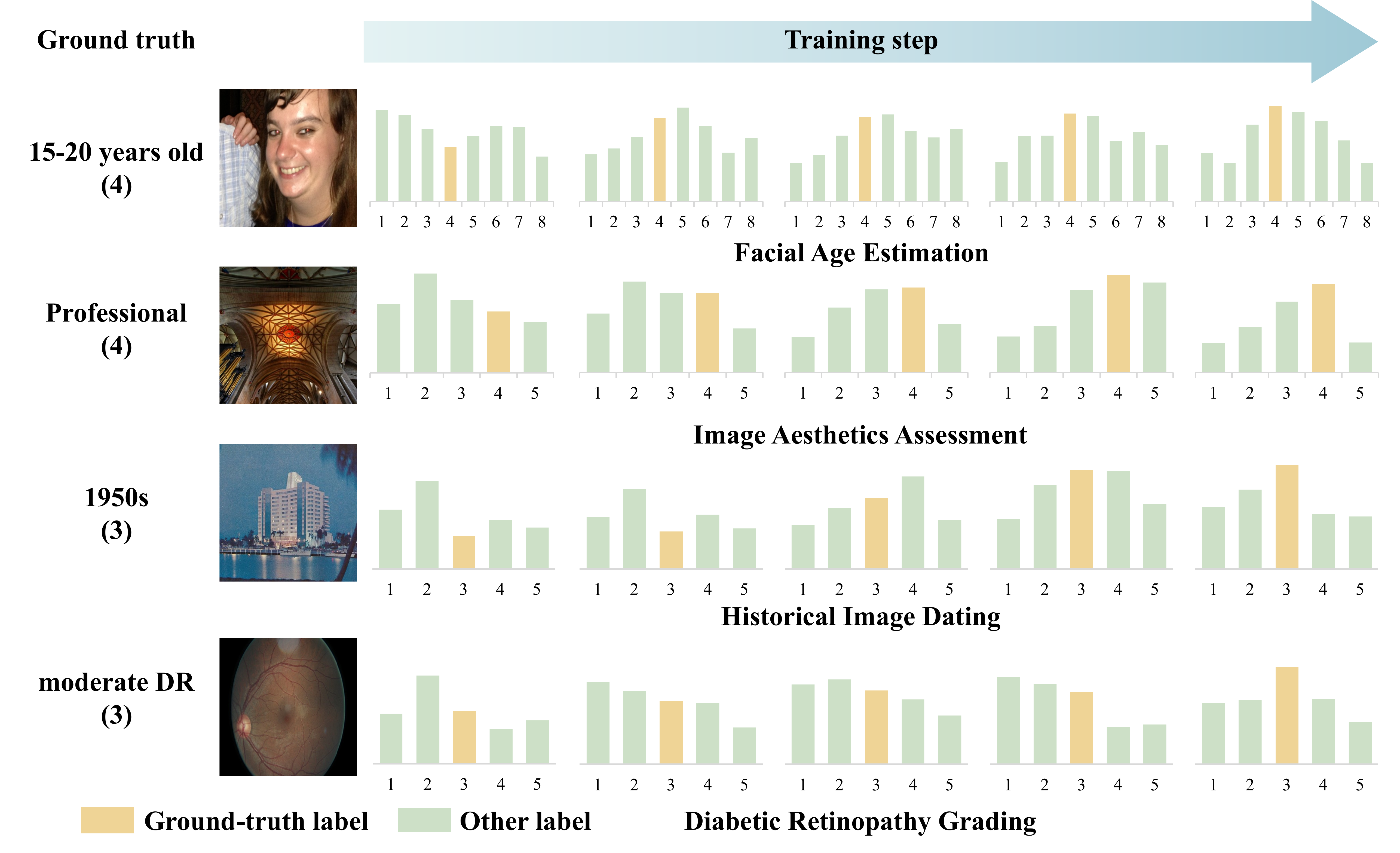}
  \caption{
   Visualization of the recovered label distributions over training steps in four scenarios.
    }
  \label{fig:vis}
\end{figure}

To gain further insight into the behavior of dynamic supervision, we visualize the evolution of the recovered label distribution during training in four scenarios, including facial age estimation, image aesthetics assessment, historical image dating and diabetic retinopathy grading.
As shown in Figure~\ref{fig:vis}, 
although the specific evolution trajectories differ across tasks, a common trend can be consistently observed: early in training, the recovered distributions remains relatively unstable and often spreads probability mass across multiple categories, whereas at later stages it becomes increasingly aligned with the underlying ordinal structure. 
This progressive refinement suggests that the model does not rely on a fixed target throughout optimization, but instead gradually recovers more reliable supervision as its representation improves.

\section{Conclusion}

In this paper, we revisited ordinal regression from the perspective of supervision design. 
We argue that static ordinal targets, as widely adopted in existing methods, may reinforce biased ordering assumptions under subjective ambiguity and class imbalance. 
To this end, we introduced \modelname, which enables training-driven supervision evolution via self-distillation.
By leveraging contrastive ordinal-aware label enhancement, \modelname recovered refined ordinal distributions that capture inter-class ambiguity and instance-level uncertainty. 
Furthermore, through CDF-based cross-layer interaction distillation, we explicitly transfered cumulative ordinal structure across network layers, ensuring consistent ordinal semantics throughout the model hierarchy.
Extensive experiments across diverse datasets demonstrated that the dynamic supervision is more effective than static ordinal encoding, especially in challenging scenarios with ambiguous or imbalanced labels. 
Our findings suggest that ordinal learning should move beyond fixed label fitting toward adaptive supervision, opening new avenues for robust representation learning under structured yet uncertain labels.


\bibliographystyle{ACM-Reference-Format}
\bibliography{sample-base}





\end{document}